\documentclass[10pt,twocolumn,letterpaper]{article}

\usepackage[pagenumbers]{cvpr} 

\definecolor{cvprblue}{rgb}{0.21,0.49,0.74}
\usepackage[pagebackref,breaklinks,colorlinks,allcolors=cvprblue]{hyperref}
\usepackage{makecell}
\usepackage{multirow}
\usepackage{pifont}    

\title{LEDiff: Latent Exposure Diffusion for HDR Generation}

\author{Chao Wang$^{1}$\quad  
Zhihao Xia$^{2}$ \quad 
Thomas Leimk\"{u}hler$^{1}$ \quad 
Karol Myszkowski$^{1}$ \quad 
Xuaner Zhang$^{2}$ \quad 
\and
$^1$MPI Informatik \qquad   $^2$Adobe
\and
{\tt\small{$\{$chaowang, karol, thomas.leimkuehler$\}$@mpi-inf.mpg.de}
} \quad
{\tt\small{$\{$cezhang, zxia$\}$@adobe.com}
}
}

\begin{document}
\twocolumn[{
  \renewcommand\twocolumn[1][]{#1}
  \maketitle
  \vspace{-6mm}
  \begin{center}
    \includegraphics[width=1.\textwidth]{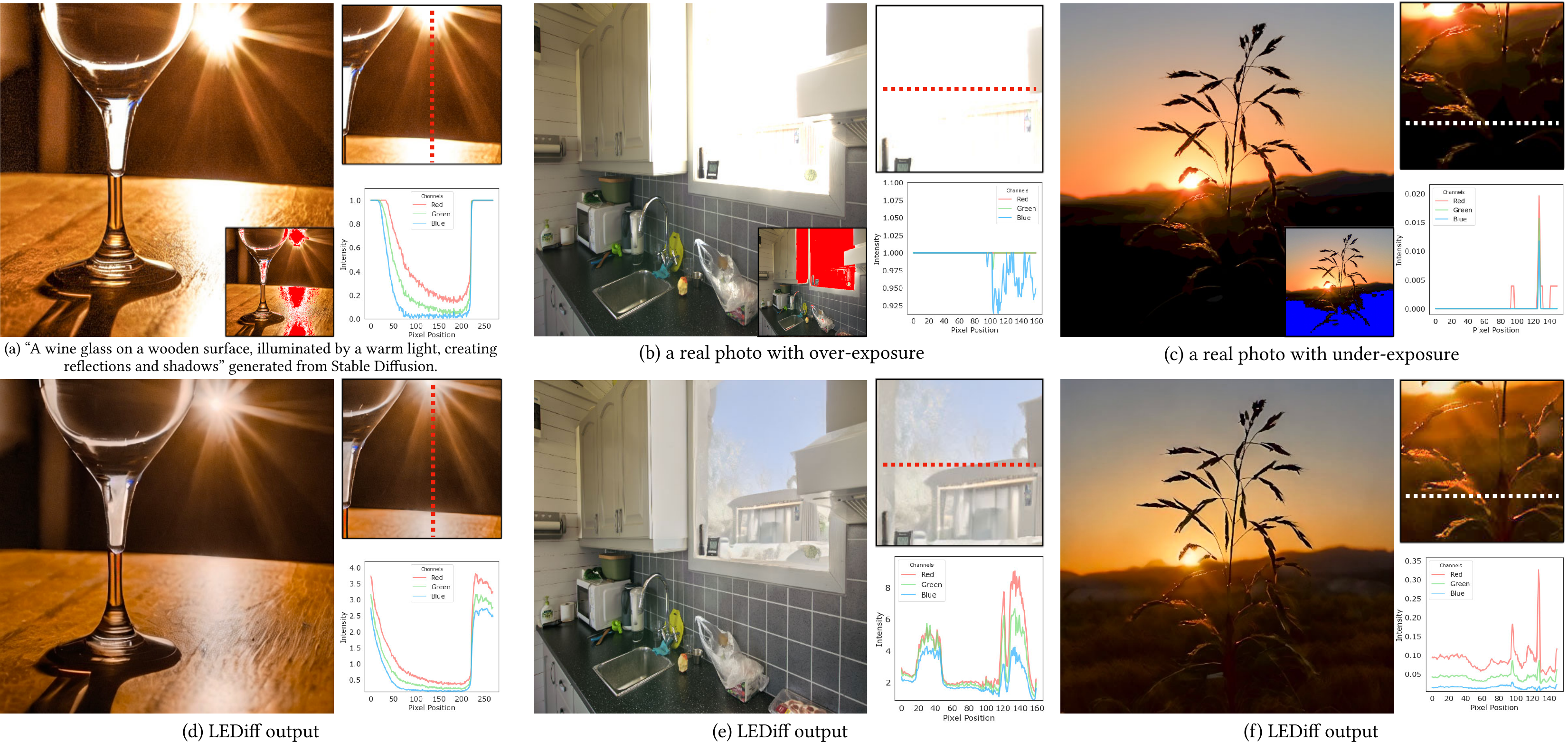}
  \end{center}
  \vspace{-5mm}
  \captionof{figure}{LEDiff enables high dynamic range (HDR) content generation with photorealistic details in both over- and under-exposed regions by performing exposure fusion in latent space, making it applicable to generated content and real photos mapped to the latent space. While existing generative models (e.g., Stable Diffusion) are restricted to low dynamic range (a) and standard cameras struggle to capture full scene dynamic range, causing clipping in highlights (b) and shadows (c), LEDiff restores both detail and dynamic range (d)--(f), as shown in scanline plots. All HDR images are tone-mapped for visualization and are best viewed on an HDR display. See the supplemental for more details.
   }
  \label{fig:teaser}
  \vspace{2mm}
}]
\maketitle

\maketitle
\vspace{-3mm}
\begin{abstract}
\label{sec:abstract}
While consumer displays increasingly support more than 10 stops of dynamic range, most image assets — such as internet photographs and generative AI content — remain limited to 8-bit low dynamic range (LDR), constraining their utility across high dynamic range (HDR) applications. Currently, no generative model can produce high-bit, high-dynamic range content in a generalizable way. Existing LDR-to-HDR conversion methods often struggle to produce photorealistic details and physically-plausible dynamic range in the clipped areas. We introduce LEDiff, a method that enables a generative model with HDR content generation through latent space fusion inspired by image-space exposure fusion techniques. It also functions as an LDR-to-HDR converter, expanding the dynamic range of existing low-dynamic range images.
Our approach uses a small HDR dataset to enable a pretrained diffusion model to recover detail and dynamic range in clipped highlights and shadows. LEDiff brings HDR capabilities to existing generative models and converts any LDR image to HDR, creating photorealistic HDR outputs for image generation, image-based lighting (HDR environment map generation), and photographic effects such as depth of field simulation, where linear HDR data is essential for realistic quality.
\end{abstract}

\section{Introduction}
\label{sec:intro}

Generative models today can produce highly realistic and creative visual content, yet they remain largely confined to 8-bit, low dynamic range (LDR) representations. This limitation causes bright highlights and deep shadows to be clipped, preventing the full capture of light and color variation present in real-world scenes. As a result, these models restrict further photographic editing, as stronger tone mapping cannot reveal any additional details. Additionally, LDR limits downstream applications, such as image-based lighting and depth of field rendering, where realistic visual quality depends on HDR data. The need for HDR content has also grown with the rise of consumer displays that support HDR, often with over 10 stops of dynamic range, making the screen feel like a window looking out at the real scene~\cite{greg_benz_hdr, eric_chan_hdr}. The longstanding popularity of LDR content — optimized for traditional display purposes through clipping, gamma correction, and quantization — may no longer hold as HDR displays become increasingly prevalent.

HDR reconstruction has been widely studied in recent years and generally falls into two main categories. The first approach focuses on multi-exposure fusion, which assumes a set of exposure-bracketed photos as input and primarily addresses alignment issues among these images~\cite{debevec2004high, sen2012robust, kalantari2017deep, wu2018deep, yan2019attention, yan2020deep, chen2023improving, liu2022ghost, tel2023alignment, niu2021hdr, xiong2021hierarchical, yan2023smae, prabhakar2021labeled, song2022selective, kong2024safnet, ye2021progressive}. The second approach centers on inverting the tone mapping applied to the input LDR image, aiming to recover missing details in clipped regions~\cite{eilertsen2017hdr, marnerides2018expandnet, liu2020single, santos2020single, endo2017deep, yu2021luminance, zhang2023revisiting, lee2018deep}. With the growth of generative models, approaches to HDR generation also emerged. GlowGAN~\cite{wang2023glowgan} is the first attempt to generate HDR content from scratch and can extend existing LDR images through GAN inversion. However, its effectiveness is limited by the class-specific training and generation constraints inherent to GANs.

Our goal is to repurpose a pre-trained latent diffusion model (e.g., Stable Diffusion~\cite{rombach2022high}) 
to generate a closely related yet distinct distribution: linear HDR content. To maintain the generative capabilities of the pre-trained model without compromising it by training on limited HDR data, we aim to retain the powerful latent space of the original latent diffusion model. 
Enabling HDR capabilities in a pre-trained generative model requires 1) hallucinating clipped area in highlights and shadows, 2) linearizing the image and expanding its dynamic range.

Interestingly, we observe that the latent space of a diffusion model strongly correlates with the image space in terms of clipping and pixel intensity; pixels that are clipped in the image space are similarly clipped in the latent space. This motivates our approach of aiming for a clipping-free latent space. We then fine-tune the decoder to linearize and expand the dynamic range, generating an HDR image in the image space.
To get a clipping-free latent code, we draw inspiration from image-space exposure fusion, applying this concept in latent space by merging latent code that represents different exposure-level captures. To generate these bracketed codes, we train a highlight generator that takes as input a latent code with highlight clipping (representing an over-exposed image) and produces a lower-exposed latent code without clipping. Similarly, we train a shadow generator that takes an under-exposed latent code as input and generates a higher-exposed latent code free from shadow clipping.
We train the highlight and shadow generators using a relatively small HDR dataset, relying on the preserved latent space to offload intensive generation tasks to the pre-trained model.
Once we have these exposure-bracketed latent codes, a learnable fusion module merges them into a single, clipping-free latent code.

In summary, our approach involves training highlight and shadow generators to hallucinate missing details in over- and under-exposed areas of the latent space and fine-tuning the decoder to linearize and expand dynamic range during decoding.

This approach offers several advantages. Leveraging the \textit{original pre-trained latent space} allows our method to act as a plug-and-play solution, converting any latent code to HDR, whether for generated content or real images encoded into the latent space. Using a \textit{learnable latent space exposure fusion}, combined with a \textit{learnable linearization decoder}, eliminates the need for hand-crafted merging features~\cite{mertens2007exposure} and avoids the necessity of estimating exposure parameters and camera response curves~\cite{bemana2024exposure} required by traditional exposure fusion algorithms.

Our method equips a generative model with HDR generation capabilities, greatly expanding its applicability across various domains. We demonstrate our method using the stable diffusion model~\cite{rombach2022high}. The main contributions of this work are as follows:
\begin{itemize}
    \item A fine-tuning approach that fully preserves a pre-trained (LDR) latent space and uses a small set of HDR data to expand dynamic range and add details to both deep shadows and bright highlights.
    \item A method that enables HDR generation in existing image and video generative models and achieves state-of-the-art performance in inverse tone mapping for real LDR images, including linearization and photorealistic reconstruction in clipped regions.
\end{itemize}

Additional results can be found on our webpage at \href{https://lediff.mpi-inf.mpg.de/}{https://lediff.mpi-inf.mpg.de/}.

\section{Related Work}
\begin{figure*}[t!]
    \centering
    \includegraphics[width=\linewidth]{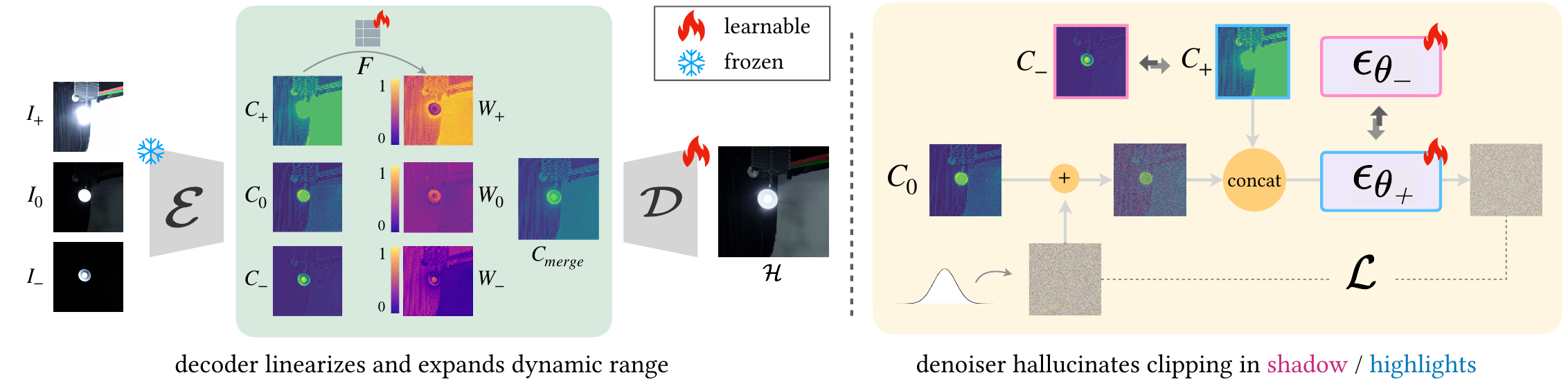}
    \vspace{-0.6cm}
    \caption{Finetuning scheme of decoder and denoiser. (Left: fine-tuning the decoder) Exposure bracketed images \(I_{+}, I_{0}, I_{-}\) are encoded via the pre-trained encoder to generate corresponding latent codes. These latent codes are fused using a learnable fusion module \(\mathcal{F}\) to produce a latent code \(\mathcal{C}_{\text{merge}}\) free of clipping, which is then decoded into an HDR image $\mathcal{H}$ through the finetuned decoder. 
    (Right: fine-tuning the denoiser) The model takes as input the latent code $\mathcal{C}_{+}$ for training highlight denoiser $\epsilon_{\theta_{-}}$ or $\mathcal{C}_{-}$ for training shadow denoiser $\epsilon_{\theta_{+}}$, along with a $\mathcal{C}_{0}$ corrupted by randomly sampled noise. 
    }
    \label{fig:Pipeline}
    \vspace{-0.5cm}
\end{figure*}

\label{sec:rw}
\subsection{Generative Image Models}
\label{sec:t2i}
The last decades have witnessed the rapid development of generative image models. In the early stage, GAN-based \cite{zhang2017stackgan, karras2019style,  xu2018attngan, zhu2018generative, zhu2019dm, li2019object, zhou2022towards, zhang2021cross, kang2023scaling, tao2023galip} methods dominate this domain and focus on domain-specific generations.

Autoregressive Transformer models, including DALL·E \cite{ramesh2021zero}, CogView \cite{ding2021cogview}, and PARTI \cite{yu2022scaling}, achieve high-quality text-to-image generation by encoding images as discrete tokens and learning text-image distributions from large datasets. However, their sequential generation process results in slower image synthesis.

Diffusion-based text-to-image (T2I) models have advanced rapidly, each iteration pushing the boundaries of efficiency and quality. GLIDE \cite{nichol2021glide} introduced Gaussian noise combined with CLIP-encoded text, enabling both classifier-guided and classifier-free T2I generation. Imagen \cite{saharia2022photorealistic} enhanced fidelity by using large language models for text encoding and scaling images through super-resolution layers. DALLE-2 \cite{ramesh2022hierarchical} linked text and image latents with a CLIP-based prior model, predicting denoised images across time steps. Stable Diffusion \cite{rombach2022high} improved efficiency by mapping images to a compact latent space, accelerating the diffusion process, while VQ-Diffusion \cite{gu2022vector} leveraged VQ-VAE \cite{van2017neural} to conduct discrete diffusion in a compressed space, producing high-quality outputs with reduced computational demands.

\subsection{High Dynamic Range Imaging}
\label{sec:hdri}
High Dynamic Range (HDR) imaging has advanced significantly over recent decades. HDR images are often reconstructed from multiple-exposure LDR images. Debevec et al. \cite{debevec2004high} achieve this by capturing static images at different exposures, estimating the camera response, and merging them. To handle motion in dynamic scenes, Sen et al. \cite{sen2012robust} introduce a patch-based method, while Kalantari et al.~\cite{kalantari2017deep} bring this into the deep learning era with a dataset for training an end-to-end HDR network. Later works build on this by using advanced architectures like U-Net \cite{wu2018deep} and transformers \cite{tel2023alignment, liu2022ghost, song2022selective, chen2023improving}, and by integrating attention, selective modules, and generative priors for enhanced HDR performance \cite{yan2019attention, yan2020deep, kong2024safnet, ye2021progressive, yan2023smae}.

An alternative approach to acquiring HDR is inverse tone mapping (ITM), where HDR content is generated from a single LDR image, addressing linearization, dynamic range extension, and hallucination \cite{banterle2017advanced}. Early ITM methods focused on linearization and dynamic range extension for display quality improvement \cite{masia2009evaluation, masia2017dynamic, akyuz2007hdr, banterle2009psychophysical, banterle2008expanding, didyk2008enhancement}. Eilertsen et al.~\cite{eilertsen2017hdr} introduced deep learning to ITM using CNNs to enhance highlight detail, inspiring later works that improve artifact reduction, linearization, and overexposure hallucination with multi-branch networks, mask mechanisms, and attention modules \cite{marnerides2018expandnet, liu2020single, santos2020single, yu2021luminance}.

Another ITM direction generates multi-exposure brackets from a single image, merging them via traditional methods. Techniques include exposure generation with 3D U-Nets \cite{endo2017deep}, recursive networks \cite{lee2018deep}, and adaptive networks for enhanced fusion \cite{zhang2023revisiting}. However, these methods often produce blurred results due to their mean-seeking nature.

More recently, Wang et al.~\cite{wang2023glowgan} modeled HDR-LDR relationships using Gaussian exposure assumptions, enabling unlabeled HDR generation and improved over-exposed region reconstruction via pre-trained models. Concurrent works \cite{bemana2024exposure, goswami2024semantic} leverage diffusion models for reconstructing HDR, creating exposure brackets without tuning, though they rely on merging to finalize HDR output. Our approach differs by performing exposure merging in the latent space, eliminating exposure parameter estimation and is able to achieve greater dynamic range.

\section{Method}



\label{sec:method}
\subsection{Preliminaries}
A latent diffusion model (LDM), such as Stable Diffusion \cite{rombach2022high}, consists of a variational autoencoder (VAE), a denoising UNet, and a text encoder. The VAE encoder $\mathcal{E}$ compresses an image into a low-dimensional latent space. The denoising UNet then iteratively refines this noisy latent code into its clean version, which is subsequently decoded by the VAE decoder $\mathcal{D}$ into an LDR image.

\begin{figure}[h!]
    \centering
    \vspace{-0.2cm}
    \includegraphics[width=\linewidth]{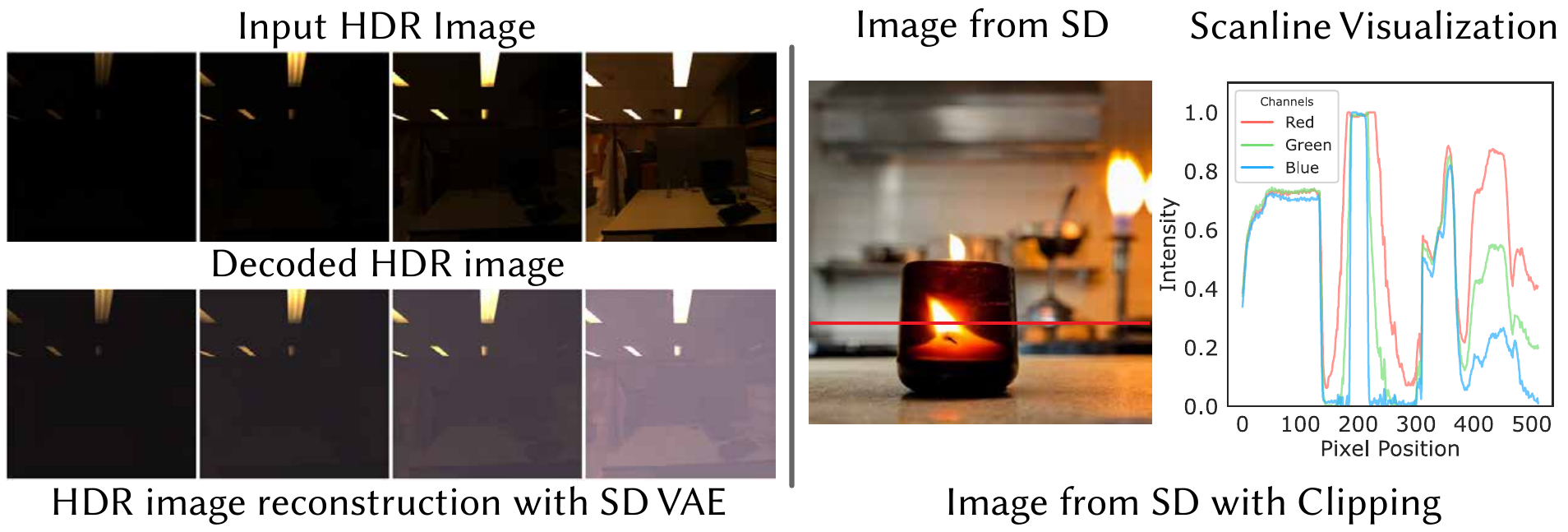}
    \caption{Limitations of the vanilla Stable Diffusion (SD) in generating HDR content. \textbf{Left:} The limitation of the SD VAE in encoding and decoding an HDR image, visualized in multiple exposure levels, which reveals a significant fidelity loss, especially in the shadow.   
    \textbf{Right:} An image generated with SD, along with a scanline that shows pixel clipping in highlight and shadow regions.}
    \label{fig:chanllenged_sd}
    \vspace{-0.2cm}
\end{figure}

LDMs are inherently limited to generating LDR images for two reasons. First, the VAE is trained exclusively on LDR images, constraining its ability to represent the dynamic range for HDR content. Second, the denoising UNet is designed to produce latent codes that encode only LDR images. These limitations are demonstrated in Fig.~\ref{fig:chanllenged_sd}. While diffusion models can plausibly hallucinate details in clipped regions — similar to their performance in inpainting tasks \cite{corneanu2024latentpaint, xie2023smartbrush, lugmayr2022repaint} — the challenge lies in guiding them to generate images with true high dynamic range and to preserve details in both highlights and shadows.

A straightforward approach to adapting LDMs for HDR generation is to fine-tune both the VAE and the UNet with HDR data. However, this requires a significant amount of HDR data and does not fully utilize the generative priors of the pre-trained model. Instead, we propose to preserve the latent space of a pre-trained latent diffusion model as is, generating LDR latent codes at multiple exposure levels — a ``latent exposure bracket'' consisting of shorter and longer exposure variants. These latent codes are then fused in the latent space using a learnable lightweight fusion module. This merged HDR latent code is subsequently decoded by a fine-tuned VAE decoder to generate an HDR image. 

In Sec.~\ref{subsec:vae}, we describe merging exposure brackets within the latent space and fine-tuning a specialized HDR decoder. In Sec.~\ref{subesec:denoiser}, we outline the approach for generating the latent exposure brackets.




\subsection{Merging Latent Exposure Brackets}
\label{subsec:vae}

A key idea in our method is to preserve the priors learned by a pre-trained latent diffusion model by restricting it to generating latent codes for LDR images -- the domain in which the model has been extensively trained on using billions of samples. The main challenge, then, is to effectively merge these LDR latent codes and decode them into an HDR image. To achieve this, we propose merging the latent LDR exposure bracket directly in the latent space and only fine-tuning the VAE decoder on a small set of HDR images to reconstruct the final HDR output, as shown in Fig. \ref{fig:Pipeline}-left.

We simulate exposure bracketed images from real HDR images. 
For each HDR image $\mathcal{H}$, we use the method from~\cite{andersson2021visualizing} to calculate the lower ($\textit{E}_{-}$) and upper ($\textit{E}_{+}$) exposure bounds needed to cover the entire scene’s dynamic range. We then project $\mathcal{H}$ into three exposure intervals, $\{\textit{E}_{-}, \textit{E}_{0}, \textit{E}_{+}\}$, where $\textit{E}_{0} = (\textit{E}_{-} + \textit{E}_{+})/2$, and randomly sample camera response curves (CRFs) from \cite{eilertsen2017hdr} to introduce non-linearity. This generates an exposure bracket of low-, medium-, and high-exposure LDR images: $I_{-}$, $I_0$, $I_{+}$, as follows:
\begin{equation}
    I_{i} = \frac{(1 + \beta) \min(\textit{E}_{i}, 1.0)^{\gamma}}{\beta + \min(\textit{E}_{i}, 1.0)^{\gamma}},
    \label{eq.crf}
\end{equation}
with ${\beta \sim \mathcal{N}(0.6, 0.1)}$ and ${\gamma \sim \mathcal{N}(0.9, 0.1)}$.
We incorporate this randomization into the CRF sampling to improve the training data diversity.
We use the pre-trained encoder $\mathcal{E}$ to map these LDR images into the latent codes: $C_{-}, C_{0}, C_{+}$.

Visualizing these latent codes (Fig. \ref{fig:Pipeline}) reveals a strong correlation between the latent space and image space, where intensity levels in over- and under-exposed regions are reflected similarly in the latent space. Thus, analogous to image-space HDR merging techniques that rely on brightness, color, and saturation to identify valid pixels \cite{mertens2007exposure}, we hypothesize that an efficient merging of the latent exposure bracket can be achieved by deriving a per-pixel weight map for each channel based on local regions of the latent codes.

To implement this, we introduce a straightforward yet effective learnable fusion module, $\mathcal{F}$. Each latent code is processed through a depth-wise convolution \cite{chollet2017xception} to produce an initial weight map, which is then normalized using a softmax function to ensure the merging weights sum to one across each pixel and channel.
This merged latent code $C_{\text{merge}}$, combining information from multiple exposures, retains scene details without clipping.
We then decode $C_{\text{merge}}$ into an HDR image in log space, fine-tuning the decoder with the original VAE's loss functions \cite{esser2021taming}, including both a reconstruction loss and a GAN loss that compares $\mathcal{D}(C_{\text{merge}})$ with $\mathcal{H}$ in log space:
\begin{equation}
\begin{aligned}
\mathcal{L}_{\mathcal{F}, \mathcal{D}} = \mathcal{L}_{\text{rec}} + \lambda \mathcal{L}_{\text{GAN}}
\end{aligned}
\end{equation}

\subsection{Generating Latent Exposure Brackets}
\label{subesec:denoiser}
With a trained $\mathcal{F}$ and a fine-tuned $\mathcal{D}$, the remaining step is to generate an exposure bracket in the latent space,  Fig. \ref{fig:Pipeline}-right.

We assume a single latent code is given, which may originate from either 1) a real image encoded into the latent space or 2) a text prompt. This allows LEDiff to naturally handle generated contents (i.e., Text-to-HDR generation) and real photos (i.e., LDR-to-HDR conversion).

When given a single LDR latent code $C$, we need to expand it into an exposure bracket $\{C_{-}, C_0, C_{+}\}$, which involves hallucinating details in over-exposed highlights and under-exposed shadows. We handle the hallucination of highlights and shadows separately but with similar processes, which is inspired by the conditioning strategy of~\cite{ke2023repurposing}. For simplicity, we describe the highlight hallucination process: generating a lower-exposure latent code, $C_-$ from $C_0$, and $C_0$ from $C_{+}$. This method implicitly learns to reduce the exposure of the conditioned image, effectively restoring details in clipped highlights.

Denote the highlight hallucination denoiser as $\epsilon_{\theta_{-}}(\cdot)$ with parameters $\theta_{-}$. We fine-tune $\epsilon_{\theta_{-}}(\cdot)$ from the pretrained Stable Diffusion denoiser to generate $C_{-}$ conditioned on the input latent $C_0$ and generate $C_0$ conditioned on the input latent $C_{+}$. The conditioning is achieved by concatenating the input condition with the noisy input before feeding it into the denoiser, with a modified first layer to handle change in channels. The rest of the network remains initialized with the pretrained weights. Fine-tuning uses a standard denoising diffusion loss.

To train $\epsilon_{\theta_{-}}(\cdot)$, we first randomly sample an exposure bracket from the HDR dataset, following the process outlined in Sec \ref{subsec:vae}. These samples are encoded using the pretrained VAE encoder, producing a latent exposure bracket $\{C_{-}, C_0, C_{+}\}$. We then corrupt the underexposed latent $C_{-}$ and $C_0$ with Gaussian noise at a randomly sampled timestamp $t$. The task of the denoising network, $\epsilon_{\theta_{-}}(\cdot)$ is to predict the noise $\hat{\epsilon}_{C_{-}, C_0} = \epsilon_{\theta_{-}}(C_0, C_{-}^{t}, t)$ and $\hat{\epsilon}_{C_0, C_{+}} = \epsilon_{\theta_{-}}(C_{+}, C_0^{t}, t)$ using the objective:
    \begin{align}
    \mathcal{L}_{C} &= \mathbb{E}_{C, \epsilon_{C_{-}, C_0} \sim \mathcal{N}(0, \mathcal{I}), t \sim \mathcal{U}(T)} \left \|  \epsilon_{C_{-}, C_0} -\hat{\epsilon}_{C_{-}, C_0} \right \|_{2}^{2}\\
    & + \mathbb{E}_{C, \epsilon_{C_0, C_{+}} \sim \mathcal{N}(0, \mathcal{I}), t \sim \mathcal{U}(T)} \left \|  \epsilon_{C_0, C_{+}} -\hat{\epsilon}_{C_0, C_{+}} \right \|_{2}^{2}
    \end{align}

During inference, given an LDR image, we iteratively apply the fine-tuned denoiser to generate its corresponding lower-exposure latent codes.

For generating a higher-exposure latent code $C_{+}$, $C_{0}$ that hallucinates shadow details, we follow the same procedure to fine-tune a conditional denoiser $\epsilon_{\theta_{+}}(\cdot)$ with parameters $\theta_{+}$, conditioned on $C_0$ and $C_{-}$ respectively.


After generating the latent exposure bracket, we merge it using $\mathcal{F}$, then decode the merged latent code into an HDR image with the fine-tuned HDR decoder.

\section{Dataset and Training}
\label{sec:experiment}
\noindent\textbf{Datasets} We gather HDR images from multiple sources \cite{liu2020single, gardner2017learning, bolduc2023beyond, kalantari2017deep, tel2023alignment, panetta2021tmo, fairchild2007hdr, polyhaven}. For the panorama datasets \cite{gardner2017learning, bolduc2023beyond, polyhaven}, we create regular images by projecting the panoramas from random camera viewpoints. In total, we obtain 36,000 HDR images, and then apply the method mentioned in Sec. \ref{subsec:vae} to generate the training pairs $\{I_{-}, I_0, I_{+}\}$ and $\{\mathcal{H}\}$.



\noindent\textbf{Training} We use this dataset to train $\mathcal{F}$ and to fine-tune $\mathcal{D}$, $\epsilon_{\theta_{-}}(\cdot)$ and $\epsilon_{\theta_{+}}(\cdot)$. $\mathcal{F}$ and $\mathcal{D}$ are trained with \(\{I_{-}, I_0, I_{+}\}\) as inputs and \(\{\mathcal{H}\}\) as the ground truth.
For $\epsilon_{\theta_{-}}(\cdot)$ training, we use the pairs \(\{I_{0}, I_{+}\}\) and \(\{I_{-}, I_0\}\), where the first image serves as the conditioning input and the second is used as the input (to be applied with noise) and ground truth. Similarly, for $\epsilon_{\theta_{+}}(\cdot)$ training, we use the pairs \(\{I_{0}, I_{-}\}\) and \(\{I_{+}, I_{0}\}\).
We fine-tune the VAE decoder for 200,000 steps with a learning rate of $10^{-6}$ and the denoiser for 400,000 steps with a learning rate of $10^{-5}$. 
All fine-tuning processes use the Adam optimizer \cite{kingma2014adam}.

\section{Applications and Evaluations}
We evaluate LEDiff across various HDR content generation tasks: text-to-image, text-to-panorama, and image-to-video (Sec. \ref{subsec:t2h}), and use it for HDR reconstruction from an LDR image (inverse tone mapping) in Sec. \ref{subsec:itm}. Our results cover diverse scenes, demonstrating the generalization ability of LEDiff and its seamless integration into applications benefiting from HDR capabilities.
\subsection{HDR Content Generation}
\label{subsec:t2h}
\begin{figure*}[h!]
    \centering
    \includegraphics[width=\linewidth]{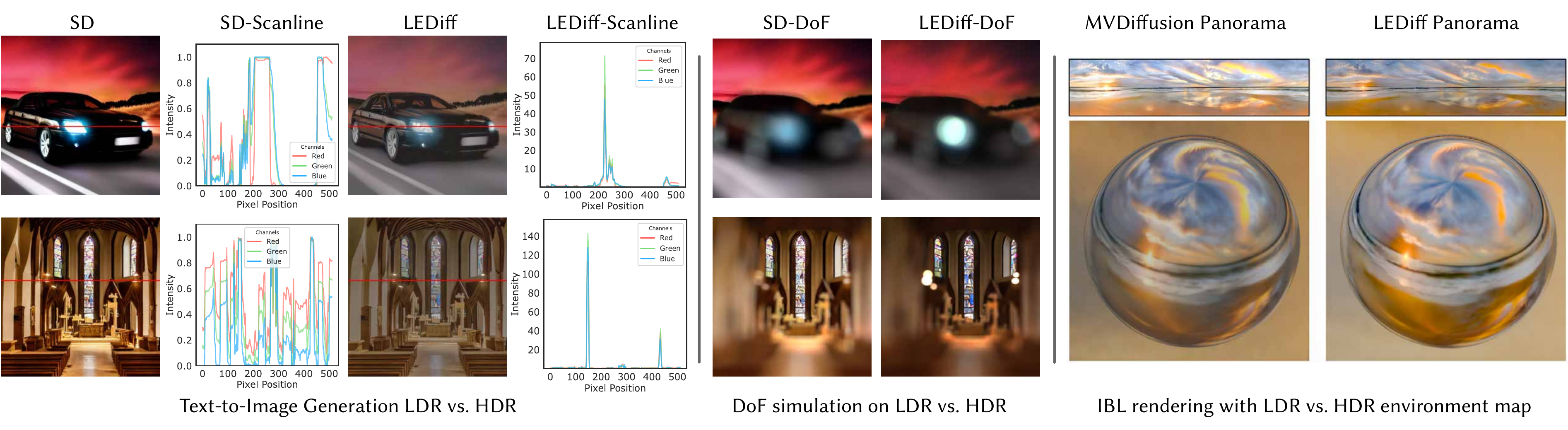}
    \vspace{-0.6cm}
    \caption{\textbf{Left:} Text to HDR image generation using prompts: (1) \textbf{``Bright car headlights on a narrow street at night''} and (2) \textbf{``A grand church interior with tall stained glass windows, intricate wooden arches, and warm lighting''}. We compare the images generated by SD and LEDiff.
    We plot a scanline passing through the car headlights and the bright luminaries to demonstrate our ability to produce a wide dynamic range with photorealistic detail. We also show an application of synthetic depth of field ($\textbf{DoF}$), 
    where a linear HDR image is crucial for rendering realistic defocus effects. \textbf{Right:} Comparisons are made between the panoramas produced by MVDiffusion \cite{Tang2023mvdiffusion} and our approach with a prompt \textbf{``A peaceful beach at sunset with soft clouds in the sky''}, along with their respective image-based lighting results. Our method leads to higher contrast and an overall more realistic appearance.}
    \label{fig:text-to-hdr}
    \vspace{-0.4cm}
\end{figure*}

\noindent\textbf{Text to HDR Image} We compare the text-to-image of LEDiff with that of Stable Diffusion (SD). Starting with a generated latent code \(C_{+}\), we apply diffusion with the denoiser \(\mathbf{\epsilon}_{\theta_{-}}\) to generate two lower exposure latent codes, \(C_{0}\) and \(C_{-}\).
The fusion module \(\mathcal{F}\) merges these latent codes into \(C_{\text{merge}} = \mathcal{F}(C_{-}, C_{0}, C_{+})\), which is then decoded into an HDR image via \(\mathcal{D}\). Fig.~\ref{fig:text-to-hdr}-left shows that LEDiff achieves a higher dynamic range without clipping, while SD output suffers from clipping. We also demonstrate the effectiveness of HDR for further image editing, such as applying synthetic defocus, where linear HDR content is crucial for realistic depth of field rendering, especially for bokeh simulation ~\cite{mildenhall2022nerf, zhang2019synthetic}.

\noindent\textbf{Text to HDR Panorama} Our method seamlessly integrates with existing SD-based panorama generation models. Using MVDiffusion \cite{Tang2023mvdiffusion} as the baseline, LEDiff enables HDR panoramas, which are essential for image-based lighting. As shown in Fig.~\ref{fig:text-to-hdr}-right, using the HDR panorama as the environment light map enhances contrast and produces more realistic highlights in the rendered image.

\noindent\textbf{Image to HDR Video} Our method can be incorporated into SD-based image-to-video models \cite{blattmann2023stable}, facilitating the generation of HDR video from a single LDR image. Examples of this application are included in the supplemental.

\begin{figure*}[h!]
    \centering
    \includegraphics[width=1.\linewidth]{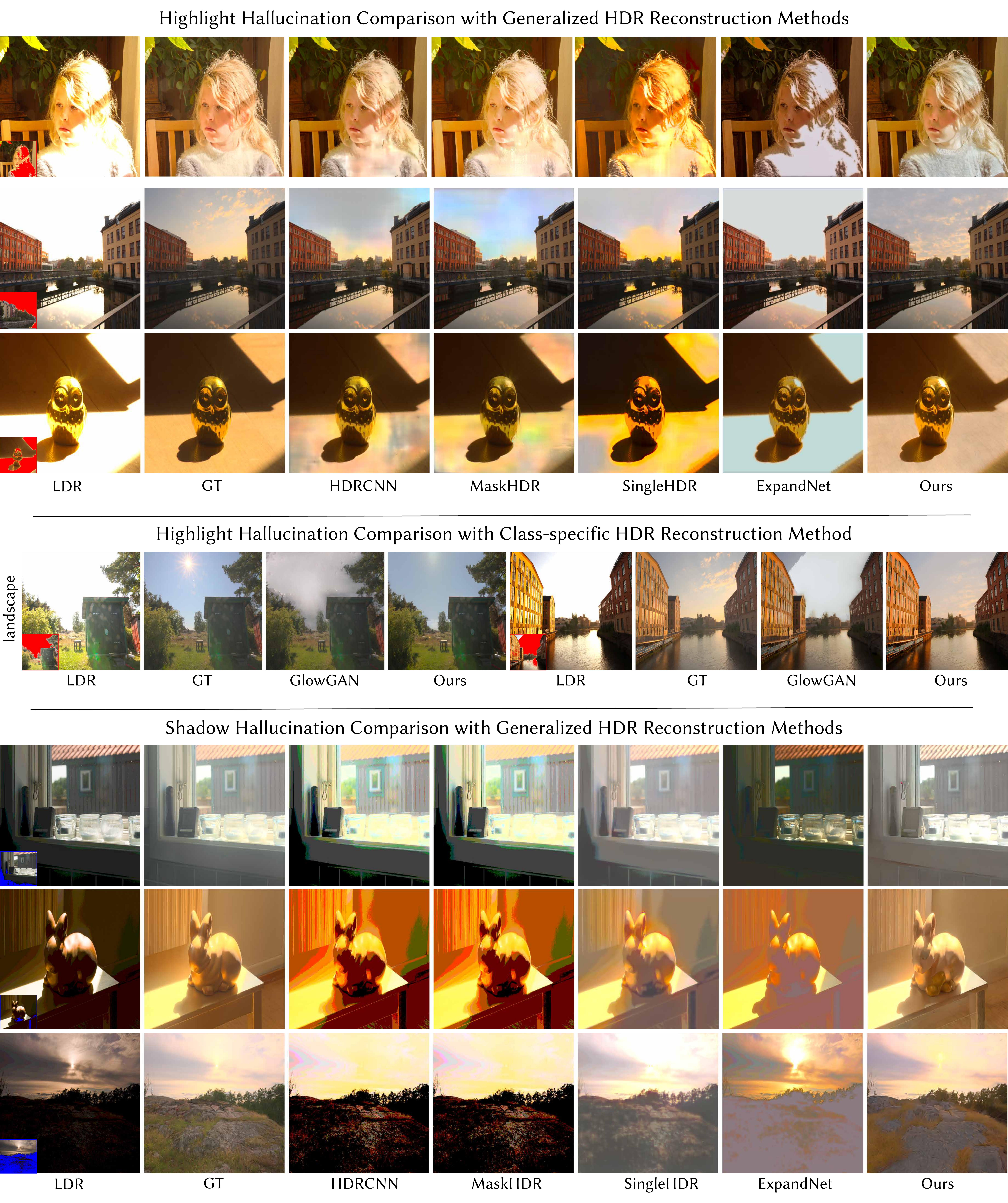}
    \vspace{-0.65cm}
    \caption{LDR-to-HDR image reconstruction comparisons.
    Our method effectively hallucinates details in both over- and under-exposed regions, while previous approaches~\cite{eilertsen2017hdr, santos2020single, liu2020single, marnerides2018expandnet, wang2023glowgan} struggle to produce plausible results, especially in shadow regions that they do not address (e.g., HDRCNN and MaskHDR yield identical results for shadow hallucination, as both methods process non-clipped regions in the same way.). Images are tone-mapped for visualization. Best viewed in HDR on an HDR display; see the supplemental for further details.
    }
    \label{fig.itm_combined}
    \vspace{-0.7cm}
\end{figure*}
\subsection{LDR to HDR Reconstruction}
\label{sec:itm}
Another promising application of LEDiff is inverse tone mapping. 
Given an LDR image \(I_{+}\) with potential clipping, we obtain its latent code \(C_{+} = \mathcal{E}(I_{+})\). Next, we generate lower exposure latents \(C_{0}\) and \(C_{-}\), which are fused into a merged latent code \(C_{\text{merge}} = \mathcal{F}(C_{-}, C_{0}, C_{+})\). Finally, the  HDR reconstruction $\hat{\mathcal{H}}$ is obtained through the decoder \(\mathcal{D}\).

We evaluate our method using standard image quality metrics. However, most established metrics 
 are optimized for LDR content. To better evaluate the HDR outputs, we also conduct a user study in a controlled environment, displaying the content on an HDR display to assess human perception of the HDR contents. 

We compare our method with five inverse tone mapping methods: \textbf{HDRCNN} \cite{eilertsen2017hdr}, \textbf{MaskHDR} \cite{santos2020single}, \textbf{SingleHDR} \cite{liu2020single}, \textbf{ExpandNet} \cite{marnerides2018expandnet}, and \textbf{GlowGAN} \cite{wang2023glowgan}. Since \textbf{GlowGAN} handles only class-specific inputs, we quantitatively evaluate it alongside our method on a subset of \textbf{SI-HDR}\cite{hanji2022comparison}, but exclude it from the user study.\\
\noindent\textbf{Image Quality Metrics}
In our quantitative evaluation, we evaluate on images from \textbf{SI-HDR} \cite{hanji2022comparison}, which are excluded from the training dataset. 
We use the full-reference HDR-VDP3 \cite{mantiuk2023hdr} and no-reference PU21-PIQE \cite{hanji2022comparison} metrics for quality evaluation. Since standard metrics struggle to assess hallucinated details in clipped regions, we also use the FID score~\cite{heusel2017gans} to evaluate image distributions. For HDR evaluation with FID, we tone-map both generated and ground-truth images to the LDR domain using Reinhard~\cite{reinhard2005dynamic}, Durand~\cite{durand2002fast}, and Liang~\cite{liang2018hybrid}. Following Chai et al.~\cite{chai2022any}, we generate 60 random $128 \times 128$ pixel crops per image, resulting in $10k$ patches for FID computation. While tone-mapped images have different feature statistics from typical LDR images, we found that our FID scores correlate well with perceived quality and maintain consistent rankings across tone-mapping methods.

Previous works \cite{eilertsen2017hdr, santos2020single, liu2020single} primarily focus on hallucinating highlight regions, with limited attention to shadows. By contrast, our approach enables effective hallucination across both highlight and shadow areas.
For highlight hallucination, we follow \cite{eilertsen2017hdr, santos2020single, wang2023glowgan} to blend the generated content with the inputs. The process models HDR luminance in the non-overexposed regions of the LDR image, aligning it with the corresponding HDR area. More details are provided in the supplemental. Fig. \ref{fig.itm_combined} shows that our method produces more natural hallucinated content within clipped highlight and shadow regions, while other methods produce low-quality artifacts and blur.
This is supported by the quantitative results in Table \ref{table:combined_evaluation}.
\begin{table*}[h!]
    \centering
    \vspace{-0.1cm}
    \small
    \scalebox{0.85}{
        \begin{tabular}{lcrrrrr|crrrrr}
            \multirow{2}{*}{Method}  & \multicolumn{5}{c}{Highlights}  & & \multicolumn{5}{c}{Shadow}  \\
             & HDR-VDP3 $\uparrow$   & PU21-PIQE $\downarrow$ &FID-R $\downarrow$ &FID-D$\downarrow$ &FID-L$\downarrow$ & & HDR-VDP3 $\uparrow$   & PU21-PIQE $\downarrow$ &FID-R $\downarrow$ &FID-D$\downarrow$ &FID-L$\downarrow$\\
            \toprule
            HDRCNN ~\cite{eilertsen2017hdr} & \cellcolor{gray!48} 6.90 $\pm$ 1.10   & 49.43 $\pm$ 6.91 & 13.39 & 16.95 & 16.67 & & 6.56 $\pm$ 1.10   & 70.95 $\pm$ 7.84 &32.55 &\cellcolor{gray!32}44.20 &26.41 \\
            MaskHDR ~\cite{santos2020single}  & \cellcolor{gray!32} 6.47 $\pm$ 1.06   & \cellcolor{gray!32} 49.38 $\pm$ 6.95 &\cellcolor{gray!32}12.83 &\cellcolor{gray!32}13.85 &\cellcolor{gray!32}15.21 & & 6.49 $\pm$ 1.24   & 70.98 $\pm$ 8.64 &\cellcolor{gray!32}32.54 & 44.43 & \cellcolor{gray!32}26.10\\
            SingleHDR ~\cite{liu2020single}  & 6.13 $\pm$ 0.87  & 49.74 $\pm$ 7.29 &28.68 & 34.53 & 29.50 & & \cellcolor{gray!48} 7.47 $\pm$ 0.78  & \cellcolor{gray!32} 46.04 $\pm$ 9.41 & 39.99 & 57.28 & 37.17\\
            ExpandNet ~\cite{marnerides2018expandnet}  & 5.23 $\pm$ 1.61   & 52.53 $\pm$ 6.50 & 18.85 & 25.49 & 21.34 & & 5.40 $\pm$ 0.93   & 60.01 $\pm$ 8.37 & 32.60 & 45.50 &29.10 \\
            Ours  & 6.16 $\pm$ 0.97   & \cellcolor{gray!48} 48.46 $\pm$ 7.04 &\cellcolor{gray!48}12.70 &\cellcolor{gray!48}13.08 &\cellcolor{gray!48}13.73& & \cellcolor{gray!32} 6.98 $\pm$ 0.83   & \cellcolor{gray!48} 43.37 $\pm$ 8.04 &\cellcolor{gray!48}20.93 &\cellcolor{gray!48}25.54 & \cellcolor{gray!48}24.26\\
            \bottomrule
            GlowGAN~\cite{wang2023glowgan}  & 5.67 $\pm$ 1.24   & 46.18 $\pm$ 8.22 & 27.44 & 40.43 & 43.85& &-- &--  &-- &-- &-- \\
            Ours$\ast$  & \underline{5.92 $\pm$ 1.17}   & \underline{45.29 $\pm$ 7.27} &\underline{15.62} &\underline{17.89} &\underline{16.19} & & --   & -- &--&-- &--\\
            \bottomrule
        \end{tabular}
    }
    \vspace{-0.1cm}
    \caption{Quantitative evaluation on content hallucination (Ours$\ast$ denotes the test results of our method on a subset of the test set, selected to meet GlowGAN's requirement for class-specific inputs). 
    Full-reference metrics are less suited to generative models due to natural deviations in hallucinated content from the original HDR scene. Our method achieves competitive performance on the full-reference HDR-VDP3 metric and outperforms alternative methods on the no-reference PU21-PIQE metric, which better captures image naturalness and perceptual fidelity. To complement these metrics, we use FID \cite{heusel2017gans} to evaluate the distribution of generated and ground-truth HDR images. Since FID is optimized for LDR, we tone map all HDR images using three tone mapping operators, resulting in three FID scores: FID-R \cite{reinhard2005dynamic}, FID-D \cite{durand2002fast}, and FID-L \cite{liang2018hybrid}. Across all tone-mapping methods, our method consistently achieves lower FID scores, demonstrating closer alignment with the HDR ground-truth distribution and superior overall quality.
   }
    \label{table:combined_evaluation}
\end{table*}
\vspace{0.2cm}
\noindent\textbf{User Study}
To comprehensively evaluate the quality of our generated results for the ITM task, we conducted a subjective study. The evaluation included 60 scenes, with half drawn from the \textbf{SIHDR} dataset and the remaining half comprising LDR images collected from online sources or real-world captures. For each scene, a competing method was randomly selected and compared to our approach, with participants tasked to express their preferences. The study was carried out using an HDR display (Dell UP3221Q, 3840×2160 resolution) in a standard office environment with natural lighting, with participants positioned 0.5 meters from the screen. The study engaged 20 participants, all possessing normal or corrected-to-normal vision, resulting in a total of 1200 pairwise comparisons. The outcome of this assessment is detailed in Table \ref{tab:user_study_results}.
The results indicate that our method significantly outperforms other approaches, demonstrating a substantial margin of superiority.
\begin{table}[h!]
\centering
\small
\scalebox{0.8}
{\begin{tabular}{ccc|ccc}
Total & Ours & HDRCNN & Total & Ours & MaskHDR \\
\midrule
310 & \cellcolor{gray!48}84.19\% & 15.81\% & 297 & \cellcolor{gray!48}88.22\% & 11.78\% \\
\midrule
Total & Ours & SingleHDR & Total & Ours & ExpandNet \\
\midrule
283 & \cellcolor{gray!48}89.40\% & 10.60\% & 310 & \cellcolor{gray!48}94.52\% & 5.48\% \\
\bottomrule
\end{tabular}}
\vspace{-0.1cm}
\caption{User study comparisons with other methods. Item \textbf{Total} shows the total number of comparisons, item \textbf{Ours}  reflects the percentage of users who preferred our results, and the other item indicates the percentage of users who chose other methods. In all cases, we observe a statistically significant preference for our method, as confirmed by a binomial test ($p < 0.01$).}
\label{tab:user_study_results}
\vspace{-0.2cm}
\end{table}

\label{subsec:itm}
\section{Ablation Study}
\label{subsec: ab}
We conduct the following experiments to illustrate the contribution of each component in our method, by excluding: 1) VAE decoder finetuning and 2) denoiser finetuning. We also use SD-based inpainting as an alternative approach for hallucinating clipped areas.
We follow Sec. \ref{sec:itm} for evaluation.
Since the inpainting needs an extra mask as input, we extract the clipped region with a threshold. 

\begin{figure}[h!]
    \centering
    \vspace{-0.35cm}
    \includegraphics[width=\linewidth]{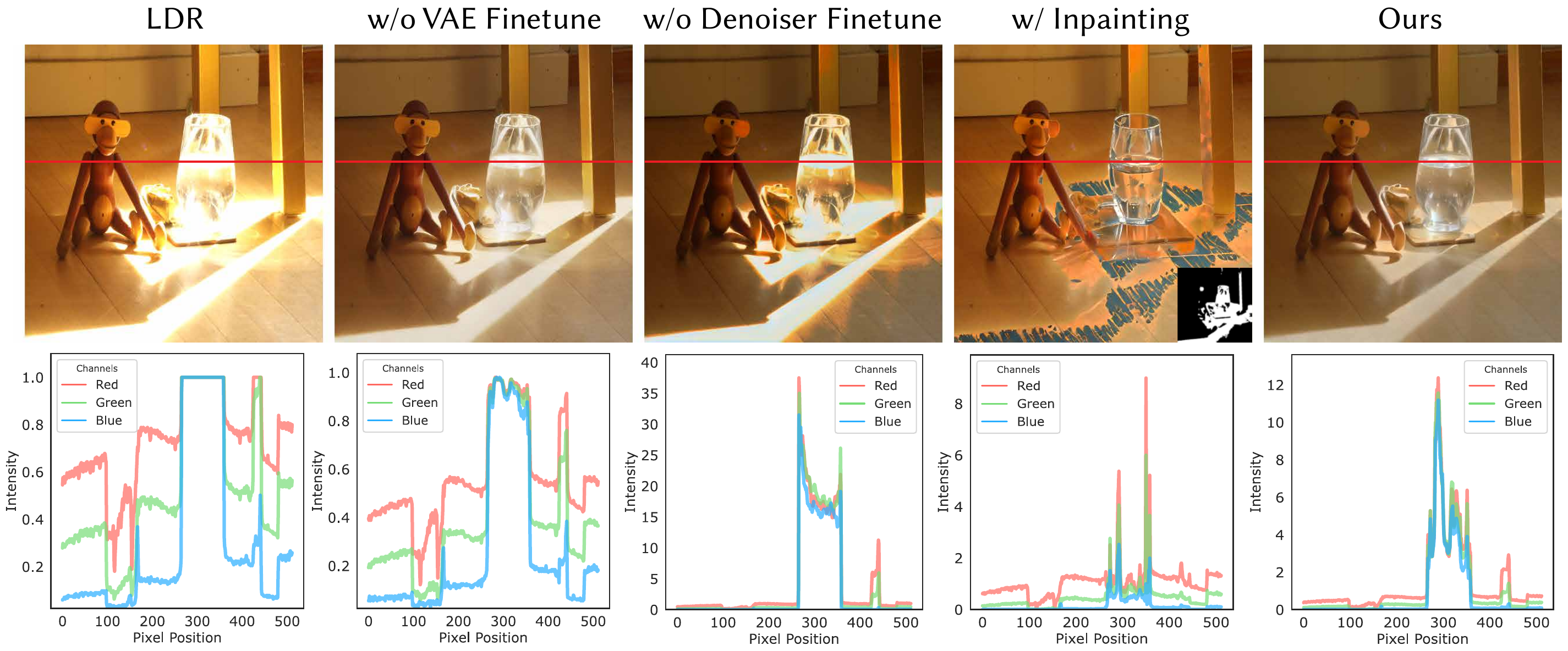}
    \vspace{-0.55cm}
    \caption{Ablation study. The first row displays the input LDR image alongside the reconstructed outputs; the second row plots the scanline. The mask used for inpainting is shown in the inset. Fine-tuning both the decoder and denoiser is essential for achieving hallucination and dynamic range extension.}
    \label{fig.ablations}
    \vspace{-0.5cm}
\end{figure}

Results are shown in Fig.~\ref{fig.ablations}, with quantitative evaluation in Table~\ref{table:ablation}. As illustrated in Fig.~\ref{fig.ablations}, omitting the decoder leads to hallucinations in overexposed regions, but restricts the content to low dynamic range (LDR). In contrast, without denoiser fine-tuning, the dynamic range is extended, but hallucinations within clipped regions are not achieved. Additionally, using an SD-based inpainting method proves suboptimal, as it relies on irregular masks that differ significantly from those seen during training, resulting in unnatural results.

We choose to use an exposure bracket with three exposure levels, as it is a common practice in image-based exposure fusion \cite{kalantari2017deep, wu2018deep, chen2023improving, liu2022ghost, song2022selective, tel2023alignment}. We explore the impact of using two and five exposure levels, detailed in the supplemental.

\begin{table}[h!]
	\centering  
\vspace{-0.2cm}
\small
\scalebox{0.85}{
    \begin{tabular}{lcrrr}
          Method  & HDR-VDP3 $\uparrow$   & PU21-PIQE $\downarrow$\\
        \toprule
         
         wo VAE &   4.67 $\pm$ 0.87   & \cellcolor{gray!32} 48.60 $\pm$ 7.02 \\ 
         
         wo denoiser  &  \cellcolor{gray!32} 5.59 $\pm$ 1.12  &  50.25 $\pm$ 6.55 \\ 
                
         inpainting &3.65 $\pm$ 1.20   & 50.18 $\pm$ 7.53\\  

         Ours  & \cellcolor{gray!48} 6.16 $\pm$ 0.97  & \cellcolor{gray!48} 48.46 $\pm$ 7.04 \\ 
         \bottomrule
    \end{tabular}
    }
    \vspace{-0.1cm}
     \caption{Ablation study on individual components. The table shows that fine-tuning both the VAE decoder and the denoiser yields the highest performance.}
\label{table:ablation}
\vspace{-0.4cm}
\end{table}

         
         
                

\section{Conclusion and Limitations}
\label{sec:conclusion}
We propose LEDiff, a method for HDR content generation that captures the full dynamic range of real-world scenes. LEDiff operates within the latent space of a pre-trained diffusion model, leveraging its generative capabilities while enabling HDR by reconstructing details in over- and under-exposed regions and performing dynamic range extension. This is accomplished by fine-tuning the VAE decoder and denoiser on a relatively small HDR dataset. LEDiff is agnostic to how the LDR latent code is generated, making it adaptable as an LDR-to-HDR converter for any LDR image encoded into latent space.
LEDiff unlocks a range of HDR applications, including HDR panorama generation for realistic image-based lighting and HDR video generation.

However, our approach has limitations. By fine-tuning Stable Diffusion, LEDiff inherits the generation limitation of SD. Additionally, we do not yet simulate degradations such as compression artifacts (e.g., ringing and blocking) or noise in the input LDR, which could improve the model’s robustness to real-world images. Addressing these limitations could improve generalization and is a promising direction for future work. 

As HDR displays become more widely available on consumer devices, we anticipate that the demand for HDR content will continue to grow, and believe that LEDiff's potential to expand HDR content creation will help meet the increasing demand for photorealistic HDR contents.

{
    \small
    \bibliographystyle{ieeenat_fullname}
    \bibliography{main}
}


\end{document}